\pdfoutput=1

\documentclass[fullpaper,final]{nldl}

\paperID{10}
\confYear{2027}
\vol{352}

\usepackage{mathtools}
\usepackage{amssymb}
\usepackage{amsthm}

\usepackage{graphicx}
\usepackage{booktabs}
\usepackage{multirow}
\usepackage{algorithm}
\usepackage{algpseudocode}
\microtypesetup{expansion=false}

\usepackage{enumitem}

\AtEveryBibitem{%
  \ifentrytype{inproceedings}{%
    \clearfield{series}%
    \clearfield{volume}%
    \clearfield{number}%
    \clearlist{publisher}%
    \clearlist{location}%
  }{}%
}
\usepackage{hyperref}
\usepackage{url}
\hypersetup{
  pdfusetitle,
  colorlinks,
  linkcolor = BrickRed,
  citecolor = NavyBlue,
  urlcolor  = Magenta!80!black,
}

\newcommand{\R}{\mathbb{R}}
\newcommand{\E}{\mathbb{E}}
\DeclareMathOperator*{\argmin}{arg\,min}

\theoremstyle{plain}

\theoremstyle{definition}
\newtheorem{observation}{Observation}

\title{Meta-LSQ: A Closed-Form Sobolev Inner Solver for Few-Shot Meta-Learning}

\author[1]{Antonin Sulc\thanks{Corresponding Author.}}
\affil[1]{Lawrence Berkeley National Laboratory}
\affil[ ]{\texttt{asulc@lbl.gov}}

\begin{document}
\maketitle

\begin{abstract}
Gradient-based meta-learners such as model-agnostic meta-learning (MAML) adapt to a new task with an inner loop of
gradient descent that has to be differentiated through, and they meta-learn every
network weight. Meta-LSQ instead fits each task with one regularised least-squares
solve over a fixed random Fourier basis and meta-learns only three scalars, a
bandwidth, a ridge and a derivative weight. Since the derivative of a sinusoid is
again a sinusoid, the same solve can also take derivative observations, and the
fitted model has exact closed-form derivatives of any order. On 10-shot sinusoid
regression Meta-LSQ trains in $14$\,s and ends with a $5.5\times$ lower error than a
first-order MAML baseline that trains $36\times$ longer, and a Gaussian-process
control suggests that most
of this gain comes from sharing one kernel across tasks. In simulated beam tuning at
DESY's ARES facility with model derivatives it is slightly more accurate than an
ALPaCA-style model and well ahead of gradient descent.
Over DESY's full range of target beams, again in simulation and from measurements
alone, its best setting has a median error of $7.3$\,\textmu m, against $14.4$ for
the published final beams of their reinforcement-learning agent, which has the
lower median on their 300 trials of small targets. On real sensor-calibration data a learned prior mean puts
it ahead of MAML on two of four targets, but an exact Gaussian process with the same
change and an ALPaCA-style model stay more accurate on most targets, and it breaks
down on non-smooth targets. On
noiseless targets outside the span of the basis, in-sample criteria such as
generalised cross-validation choose too small a ridge.
\end{abstract}

\section{Introduction}
\label{sec:intro}

\begin{figure*}[tb]
\centering
\includegraphics[width=\textwidth]{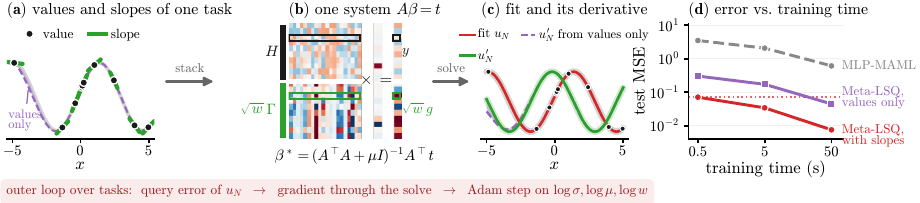}
\caption{\textbf{Meta-LSQ on one held-out sinusoid task.} \textbf{(a)}~The
support set gives ten values and, at the same points, ten slopes. A value fixes a
point of the curve and a slope fixes its direction there. Meta-LSQ fitted to the
values alone (dashed) passes through every point but drifts from the truth where
the points are sparse (shaded).
\textbf{(b)}~Each value is a row of $H$ with target $y_i$ and each slope a row of
$\Gamma$ with target $g_i$, in the same unknowns $\beta$; the two rows and targets of
one point are outlined. A slope row is the value row shifted by $\pi/2$ and scaled by
$W_j$ (20 of $N{=}200$ features shown). \textbf{(c)}~The fitted $u_N$ and its closed-form
derivative $u_N'$ over the true task and its derivative (pale). The values-only fit
of (a) also has a closed-form derivative (dashed), but it is wrong where the points
are sparse; the slope rows correct it. Only $\sigma$, $\mu$ and $w$ are meta-learned, by the outer loop
shown at the bottom. \textbf{(d)}~Test MSE against training time.
Meta-LSQ-Sobolev after $0.5$\,s (dotted) is below MLP-MAML after $50$\,s
(Table~\ref{tab:wallclock}).}
\label{fig:teaser}
\end{figure*}

In the classical pipeline a network is trained on each task separately, from a
random start and by many steps of gradient descent, which needs far more than a
handful of examples. A meta-learner is instead trained on many related tasks so that
a new task can be fitted from a few examples. Model-Agnostic Meta-Learning
(MAML)~\citep{finn2017maml}, the most widely used recipe, still fits each task by
gradient descent, but only for a few steps from a meta-learned start. An inner loop
takes $T$ gradient steps on a task's support set, giving adapted weights
$\theta_\tau = \theta_0 - \eta \sum_{t<T} \nabla_\theta
\mathcal{L}^s_\tau(\theta_t)$, and an outer loop tunes the initialisation
$\theta_0$ so that the query loss after adaptation is small. The outer gradient has
to differentiate through the $T$ inner updates, kept on the graph, which needs
double backpropagation. First-order variants~\citep{nichol2018reptile} drop the
second-order terms, and \citet{jamal2021lazyhessian} shorten the meta-gradient
graph for long inner loops with a lazy teacher--student scheme. In all of these
the meta-learned object is still the whole initialisation, so the outer search
runs over as many dimensions as the network has weights, typically thousands.

If the inner loop is replaced by a convex solver, the iterative fit goes away and
so does part of the second-order cost. ALPaCA~\citep{harrison2018alpaca} does
closed-form Bayesian linear regression with a meta-learned prior on top of a
learned encoder, and was evaluated on the same sinusoid benchmark we use;
R2D2~\citep{bertinetto2019r2d2} puts ridge regression on a convolutional
backbone, and MetaOptNet~\citep{lee2019metaoptnet} a convex SVM. The inner solve
is one-shot and differentiable in all three, but the features still come from a
deep network that is meta-learned along with it, so the number of meta-parameters
does not go down.

\paragraph{Approach.}
We keep the closed-form solve but drop the learned feature extractor, so Meta-LSQ
runs no gradient descent on a task at all. Each task is fitted by one regularised
least-squares solve over a frozen random Fourier basis, the outer loop
differentiates through that solve instead of through gradient steps, and only two or
three scalars describing the basis are meta-learned.\footnote{The code and the stored results behind every number and
figure are at \url{https://github.com/sulcantonin/metalsq-public}.} Fixing the
features costs expressivity, and the method is
meant for smooth, low-dimensional families that the basis represents well
(failures outside it are in Sec.~\ref{sec:exp-limits}).

The solve can also use derivative observations at little cost. The derivative
of a sine feature is the same feature shifted by a quarter period
(Fig.~\ref{fig:teaser}b), so a derivative constraint is linear in the same
coefficients as a value constraint. Both go into one linear system, which we call
\emph{Sobolev fitting}. Fitting derivatives with an MLP (Sobolev
training~\citep{czarnecki2017sobolev}) requires backpropagating through a backward
pass, which roughly doubles graph depth and memory; in Meta-LSQ it adds $Md$ rows to
the linear system.

\paragraph{Contributions.}
\begin{itemize}[leftmargin=*,itemsep=1pt,topsep=2pt]
\item Meta-LSQ (Sec.~\ref{sec:method}). It replaces MAML's inner loop with one
least-squares solve on values and derivatives, and the learned initialisation with
two or three scalars. The meta-gradient comes exactly from the solve, with no
unrolling. The solve approximates gradient-enhanced kriging, so each scalar is a
kernel length-scale or a noise ratio.
\item An analysis of the derivative rows (Sec.~\ref{sec:mechanism}). With ten
support points, the ten slopes add only about three new independent directions to
the ten values, but they lower the ratio of the largest to the tenth singular
value more than a hundredfold. The individual coefficients stay unidentified, but
on held-out points the fitted function's values and derivatives become three to five
times more accurate.
\item Experiments (Sec.~\ref{sec:experiments}) against MAML (a strong first-order
variant, a matched training time, and MAML given the same derivatives), an
ALPaCA-style model and gradient-enhanced kriging; on real sensor-calibration data;
in a simulated beam-tuning loop; and on the cases where the method fails.
\item A limit of standard ridge selection (Sec.~\ref{sec:exp-ridge}). The ridge
$\mu$ decides how far the fit may stray away from the data. With fewer observations
than features and no noise, a tiny ridge fits the support points exactly, so methods
that choose $\mu$ by how well those points are fitted, such as generalised
cross-validation, can prefer the smallest $\mu$ even when the fit is far off
elsewhere. The ridge then has to be set from the noise level instead. The
meta-learned ridge does follow the noise level, and we give a rule for a minimum
ridge in deployment.
\end{itemize}

\section{Related Work}
\label{sec:related}

\paragraph{Closed-form inner solvers.}
The closest prior method is ALPaCA~\citep{harrison2018alpaca}, whose learned
encoder Meta-LSQ replaces with frozen features; R2D2 and MetaOptNet also learn
their backbone (Sec.~\ref{sec:intro}). iMAML~\citep{rajeswaran2019imaml} gets the outer gradient from the
implicit function theorem instead of unrolling, and \citet{lorraine2020implicithg}
approximate hypergradients the same way; here the inner optimum is the ridge
solution, so they are exact up to round-off.
Conditional Neural Processes~\citep{garnelo2018cnp} amortise the support-to-query
mapping through a learned set-encoder, and Deep Kernel
Transfer~\citep{patacchiola2020dkt} learns a Gaussian-process (GP) kernel. Meta-LSQ learns no
backbone or encoder, only three scalars of a fixed kernel family.

\paragraph{Derivative-augmented fitting and random features.}
Sobolev training~\citep{czarnecki2017sobolev} fits derivatives at
double-backpropagation cost, and gradient-enhanced kriging
(GEK)~\citep{laurent2019gek} does so in GP regression at $\mathcal{O}(M^3 (d{+}1)^3)$, after Hermite RBF
interpolation~\citep{wu1992hermite} and kriging with
derivatives~\citep{morris1993derivatives}. Random features are due to
\citet{rahimi2007rff}, \citet{tancik2020fourierfeat} tied them to spectral bias,
and extreme learning machines~\citep{huang2015elm} and
PIELM~\citep{dwivedi2020pielm} fit closed-form read-outs over a frozen random
basis. FastLSQ~\citep{sulc2026fastlsq} solves one PDE at a time with sine
features, and physics-informed networks~\citep{raissi2019pinn} are retrained per
instance; meta-learning the spectrum of such a fit across a family is, as far as
we know, new.

\section{Background}
\label{sec:background}

\paragraph{Random Fourier features.}
A smooth function can be written as a sum of waves, and if the frequencies of the
waves are fixed in advance, only their amplitudes are left to fit, which is a
linear problem. Random Fourier features~\citep{rahimi2007rff} work this way. We
draw $N$ frequencies $W_j$ at random, give each a random phase
$b_j \sim \mathcal{U}[0,2\pi)$, and model the function as
\begin{equation}
\label{eq:rff-basis}
u_N(x) = \frac{1}{\sqrt{N}} \sum_{j=1}^{N} \beta_j \sin(W_j^\top x + b_j).
\end{equation}
The waves are drawn once and frozen, and only the amplitudes $\beta \in \R^N$ are
ever fitted. How smooth the fits are is decided by the spread of the frequencies.
With $W_j \sim \mathcal{N}(0,\sigma^2 I)$, a small bandwidth $\sigma$ gives slow
waves and smooth fits, while a large one gives fast waves that can follow fine
detail but need more data to pin down. By Bochner's theorem~\citep{bochner1933},
for large $N$ fitting $\beta$ behaves like kernel regression with a Gaussian kernel
of length-scale $1/\sigma$, so the usual intuitions about kernel bandwidths carry
over. The model is nonlinear in $x$ but linear in $\beta$, so the inner loop can be
solved in closed form.

\paragraph{Closed-form derivatives of the basis.}
Differentiating a sine shifts its phase by a quarter period and multiplies it by the
frequency, $\tfrac{d}{dx}\sin(Wx+b) = W\sin(Wx+b+\pi/2)$. Repeating this, a derivative of
any order, written with a multi-index $\alpha$ in $d$ dimensions, is
\begin{equation}
\label{eq:cyclic}
D^\alpha \sin z_j = W_j^{\alpha}\, \sin\big(z_j + |\alpha|\,\tfrac{\pi}{2}\big),
\end{equation}
where $z_j = W_j^\top x + b_j$ and $W_j^{\alpha} = \prod_k W_{kj}^{\alpha_k}$, so
every derivative costs one sine evaluation and is exact up to round-off, without
the step-size error of finite differences. Its size grows with the factor
$W_j^{\alpha}$, so the bandwidth sets how large high orders get. FastLSQ~\citep{sulc2026fastlsq} builds a calculus of
differential and integral operators on this identity; we only need these two
facts.

\section{Meta-LSQ}
\label{sec:method}

\paragraph{Scope and assumptions.}
We assume throughout that \textbf{(A1)} the task functions are smooth and
\textbf{(A2)} the support values are clean or nearly so. When, in addition, the
functions lie close to the span of the basis, as the sinusoids do by construction,
the best ridge is close to zero and the solve is almost exact, and this is the
setting of our main results. Fitting derivatives as well as values
needs derivative observations at the support points. Analytic generators,
simulators and physical models usually provide them with the values (as a
sensitivity, a force or a residual); when none are available we fit the values
alone, as we do on the real data (Sec.~\ref{sec:exp-real}).

\subsection{Adaptation as a Single Least-Squares Solve}
\label{sec:sobolev-solve}

A task gives us $M$ support points $x_i$, the value $y_i$ at each and, when
available, the gradient $g_i \in \R^d$. Write the model~\eqref{eq:rff-basis} as
$u_N(x) = h(x)^\top\beta$ with $h_j(x) = \sin(W_j^\top x + b_j)/\sqrt{N}$.
By~\eqref{eq:cyclic} its partial derivatives are linear in the same $\beta$,
\begin{equation}
\label{eq:grad-features}
\partial_k u_N(x) = \gamma_k(x)^\top\beta,
\end{equation}
with $\gamma_{k,j}(x) = W_{kj}\cos(W_j^\top x + b_j)/\sqrt{N}$,
so a measured value and a measured slope at $x_i$ are two linear equations in the
same unknowns,
\begin{equation}
\label{eq:rows}
h(x_i)^\top\beta = y_i, \quad \gamma_k(x_i)^\top\beta = g_{ik}\ \ (k \le d).
\end{equation}
With the $h(x_i)^\top$ as the rows of $H$, the $\gamma_k(x_i)^\top$ for all $i$ and
$k$ as the rows of $\Gamma$, and the $g_{ik}$ in $g$ in the same order, we stack
them into one design matrix and one target vector,
\begin{equation}
\label{eq:stack}
A = \begin{bmatrix} H \\ \sqrt{w}\,\Gamma \end{bmatrix},\qquad
t = \begin{bmatrix} y \\ \sqrt{w}\,g \end{bmatrix},
\end{equation}
where the weight $w$ says how much a slope counts relative to a value. Adaptation is
then a single ridge solve,
\begin{align}
\label{eq:beta-star}
\beta^\ast(\sigma,\mu,w)
  &= \argmin_\beta \|A\beta - t\|_2^2 + \mu\|\beta\|_2^2 \nonumber\\
  &= (A^\top\! A + \mu I)^{-1} A^\top t ,
\end{align}
where the ridge $\mu$ keeps the amplitudes small in directions the data say
nothing about, and $\sigma$ enters through the frequencies $W = \sigma\hat W$.
With $w = 0$ this is plain ridge regression on the values, and with derivatives the
system only gets more rows, $M(d{+}1)$ instead of $M$.

The derivatives of the fit, $\gamma_k(x)^\top\beta^\ast$, are available in closed
form either way. The slope rows matter when slopes are measured. A fit to the values
alone passes through every $y_i$ but can have the wrong slope at and between the
points (Fig.~\ref{fig:teaser}a), and the rows in~\eqref{eq:rows} rule such fits out
at the cost of a taller $A$ (Sec.~\ref{sec:mechanism} measures what they add). We call the method with
derivative rows \emph{Meta-LSQ-Sobolev} and the one with $w{=}0$ \emph{value-only
Meta-LSQ}.

\paragraph{Kernel view.}
\label{sec:dictionary}
For $N\to\infty$ the solve tends to kernel ridge regression with an RBF kernel of
length-scale $1/\sigma$ and, with $w>0$, to the posterior mean of a GP that also
observes derivatives, that is, to GEK. With a Gaussian prior of variance $\sigma_p^2$ on
$\beta$ and Gaussian noise of variance $\sigma_y^2$ on the values and $\sigma_g^2$
on the derivatives, \eqref{eq:beta-star} is the most probable $\beta$ when $\mu =
\sigma_y^2/\sigma_p^2$ and $w = \sigma_y^2/\sigma_g^2$ (App.~\ref{app:posterior}).
Meta-LSQ thus learns one kernel for the whole family, and the two ratios predict
that $\mu$ should grow with value noise and $w$ should fall with derivative noise,
which we test in Secs.~\ref{sec:exp-mechanism} and~\ref{sec:exp-limits}. The same
model gives error bars on any value or derivative of the fit, but with a residual
plug-in noise estimate the nominal $90\%$ intervals cover only $31$--$64\%$, so we
do not treat them as calibrated (App.~\ref{app:calib}).

\subsection{The Outer Objective}
\label{sec:outer-loop}

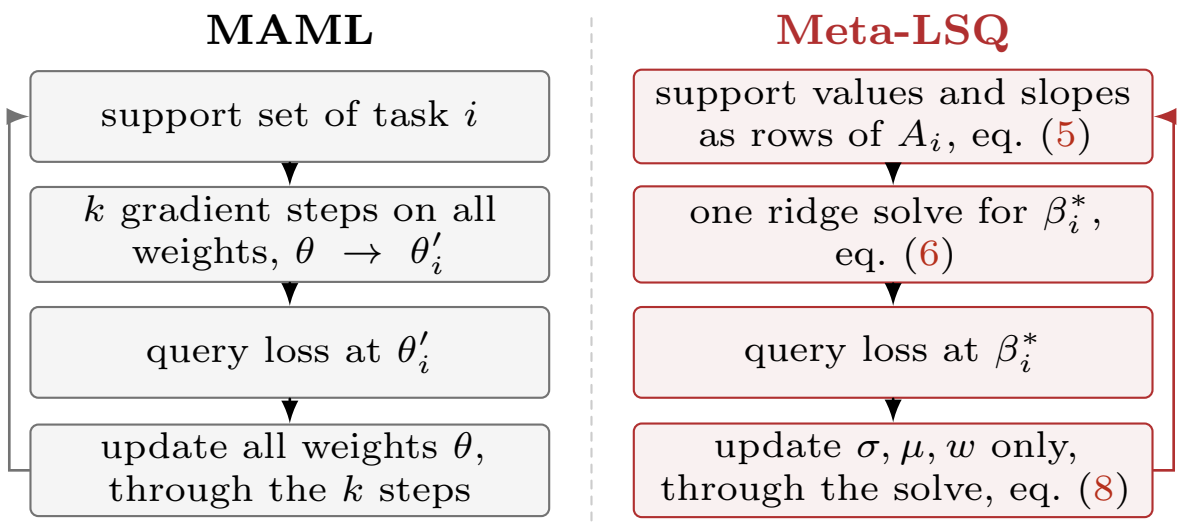
\begin{figure}[tb]
\centering
\begin{tikzpicture}[x=\linewidth, y=1cm, line cap=round,
    box/.style={draw=black!55, fill=black!4, rounded corners=1.5pt, line width=0.4pt,
                align=center, text width=0.425\linewidth, inner sep=1.5pt, minimum height=0.6cm,
                font=\scriptsize},
    lsq/.style={box, draw=wfred, fill=wfred!7},
    arr/.style={-latex, line width=0.5pt}]
\definecolor{wfred}{HTML}{B03030}
\node[font=\footnotesize\bfseries, anchor=base] at (0.245, 0.48) {MAML};
\node[font=\footnotesize\bfseries, anchor=base, text=wfred] at (0.755, 0.48) {Meta-LSQ};
\node[box] (m1) at (0.245, 0)     {support set of task $i$};
\node[box] (m2) at (0.245, -0.78) {$k$ gradient steps on all\\weights, $\theta \to \theta_i'$};
\node[box] (m3) at (0.245, -1.56) {query loss at $\theta_i'$};
\node[box] (m4) at (0.245, -2.34) {update all weights $\theta$,\\through the $k$ steps};
\node[lsq] (l1) at (0.755, 0)     {support values and slopes\\as rows of $A_i$, eq.~\eqref{eq:stack}};
\node[lsq] (l2) at (0.755, -0.78) {one ridge solve for $\beta_i^\ast$,\\eq.~\eqref{eq:beta-star}};
\node[lsq] (l3) at (0.755, -1.56) {query loss at $\beta_i^\ast$};
\node[lsq] (l4) at (0.755, -2.34) {update $\sigma, \mu, w$ only,\\through the solve, eq.~\eqref{eq:ift}};
\foreach \a/\b in {m1/m2, m2/m3, m3/m4, l1/l2, l2/l3, l3/l4} \draw[arr] (\a) -- (\b);
\draw[arr, black!55] (m4.west) -- ++(-0.017, 0) |- (m1.west);
\draw[arr, wfred] (l4.east) -- ++(0.017, 0) |- (l1.east);
\draw[black!20, line width=0.4pt, dash pattern=on 1.5pt off 1.5pt] (0.5, 0.65) -- (0.5, -2.7);
\end{tikzpicture}
\caption{One meta-training step. MAML adapts with gradient steps on all weights and
differentiates through them; Meta-LSQ adapts with one ridge solve over frozen
features, differentiates through it by eq.~\eqref{eq:ift} and learns three scalars.}
\label{fig:workflow}
\end{figure}

The three scalars should be chosen so that a single solve generalises across the
family, so each training task is fitted on its support set and the fit is scored
on the task's $K_q$ query points,
\begin{equation}
\label{eq:meta-objective}
\min_{\sigma,\mu,w > 0}\;\;
\E_{\tau}\big[\tfrac{1}{K_q}\|H^q_\tau(\sigma)\,\beta^\ast_\tau(\sigma,\mu,w) - y^q_\tau\|_2^2\big],
\end{equation}
in which only the three positive scalars are learned (two for value-only
Meta-LSQ). The frozen draw $(\hat W, b)$ and the number of
features $N$ are set once at initialisation. $N$ plays a role similar to the width
of an MLP, and the test error hardly depends on it (App.~\ref{app:details}). We learn the logarithms of the scalars, which keeps
them positive, rescale the frozen draw as $W = \sigma\hat W$ at every outer step,
and use Adam (Fig.~\ref{fig:workflow}; Algorithm~\ref{alg:meta-lsq} in
App.~\ref{app:alg}).

\paragraph{Gradient through the solve.}
Write $R = A^\top\! A + \mu I$, so that $R\beta^\ast = A^\top t$. Differentiating
both sides with respect to a scalar $\theta \in \{\sigma, \mu, w\}$ (the implicit
function theorem) gives
\begin{equation}
\label{eq:ift}
\partial_\theta \beta^\ast = R^{-1}\big(\partial_\theta (A^\top t) - (\partial_\theta R)\,\beta^\ast\big),
\end{equation}
for the ridge simply $\partial_\mu \beta^\ast = -R^{-1}\beta^\ast$. The query loss
$\mathcal{L}$ of a task then has $\partial_\theta \mathcal{L} = (\nabla_{\beta^\ast}
\mathcal{L})^\top \partial_\theta \beta^\ast$ (plus, for $\sigma$, a direct term
through $H^q$). $R^{-1}$ reuses the forward factorisation and nothing is unrolled.

\paragraph{Cost of an outer step.}
The solve costs $\mathcal{O}(M(d{+}1)N^2 + N^3)$, under a millisecond per task here.
An outer step costs about as much as a MAML step, and Meta-LSQ trains faster in
Table~\ref{tab:maml} because it needs a quarter as many steps (App.~\ref{app:details}).

\subsection{Rank and Conditioning of the Stacked Design}
\label{sec:mechanism}

Ten slopes can help a fit with 200 unknowns by adding new independent directions to
the design or by making the system better conditioned, and we measure both.

\begin{observation}[Slope rows add few new directions]
\label{obs:rank}
In exact arithmetic, $M(d{+}1) \leq N$ value and slope rows at distinct points are
almost surely linearly independent, so the rank could grow from $M$ to $M(d{+}1)$.
Numerically, what counts is the effective rank, the number of singular values above
a relative tolerance. At a small bandwidth the features are smooth, a slope is
almost determined by nearby values, and the effective rank stays well below
$M(d{+}1)$.
\end{observation}

We measured both with ten support points at the learned bandwidth
(Fig.~\ref{fig:mechanism}c). The ten value rows are essentially independent. The ten
slope rows add only about three more independent directions (relative tolerance
$10^{-6}$), but they lower the
ratio of the largest to the tenth singular value from $10^{4.7}$ to $10^{2.3}$.
The full stacked design stays numerically singular, and neither design determines
the 200 coefficients, yet with the slope rows the fitted function is about five
times more accurate on random targets in the span of the basis, for its values and
its derivatives (Fig.~\ref{fig:mechanism}d), and three to four times on the sinusoid
tasks (App.~\ref{app:details}).

\section{Experiments}
\label{sec:experiments}

All experiments run single-threaded on CPU, and we only compare times within one
experiment. Unless we say otherwise we report the mean over 5 seeds, where a seed
redraws the frozen basis and the training and test tasks (further protocol details
are in App.~\ref{app:details}).

\subsection{Accuracy and Training Cost}
\label{sec:exp-works}

\paragraph{Setup.} We use the usual sinusoid family $y = a\sin(x+\phi)$ with
$a\sim\mathcal{U}[0.1,5]$, $\phi\sim\mathcal{U}[0,\pi]$ and $x \in [-5,5]$, with
$K{=}10$ support and $K_q{=}50$ query points at test time and 200 held-out tasks.
We compare with a \emph{standard MLP-MAML} (two hidden layers of 40 units, one inner
step, 8000 iterations) and a \emph{strong MLP-MAML}, a larger first-order network with
five inner steps trained for up to 60k iterations. We also run MLP-MAML with a
Sobolev loss on the support gradients (\emph{MAML\,$+$\,Sob.}), an
\emph{ALPaCA-style} model and GEK. \emph{Meta-LSQ} has $N{=}200$ frozen features
and is trained for 2000 iterations; Meta-LSQ-Sobolev is also given the support
gradients (App.~\ref{app:details}).

\begin{table}[tb]
\centering
\caption{10-shot sinusoid regression, 200 test tasks, meta-iterations in
parentheses. Mean\,$\pm$\,std over 5 seeds (3 for ALPaCA-style and MAML\,$+$\,Sob.;
one run per budget for strong MAML). ``learn.'': learnable parameters; ``ms/t'':
per-task adaptation time. GEK uses an RBF kernel, $^\ddagger$fitted per task by
marginal likelihood or $^\ast$shared and meta-learned like Meta-LSQ's scalars
(App.~\ref{app:gp}). Bold: best without per-task search.}
\label{tab:maml}
\footnotesize
\setlength{\tabcolsep}{2pt}
\begin{tabular}{lrrrl}
\toprule
method & learn. & train\,s & ms/t & test MSE \\
\midrule
MAML std.\ (8k)       & 1\,761 & 51  & $0.5$ & $0.70 \pm 0.13$ \\
MAML strong (10k)     & 8\,513 & 130 & $1.8$ & $0.328$ \\
\quad'' (30k)         &        & 399 &        & $0.144$ \\
\quad'' (60k)         &        & 512 &        & $0.104$ \\
MAML $+$ Sob.\ (8k)   & 1\,761 & 78  & $1.0$ & $0.216 \pm 0.031$ \\
ALPaCA-style (8k)     & 1\,721 & 12  & n/a   & $0.026 \pm 0.017$ \\
GEK, per task$^\ddagger$ & n/a & n/a & $2.9$ & $0.013 \pm 0.008$ \\
GEK, shared (2k)$^\ast$ & 3 & n/a & n/a & $0.026 \pm 0.014$ \\
\midrule
Meta-LSQ $w{=}0$ (2k) & 2      & 11  & $0.4$ & $0.060 \pm 0.043$ \\
Meta-LSQ-Sob.\ (2k)   & 3      & 14  & $0.9$ & $\mathbf{0.019 \pm 0.016}$ \\
\bottomrule
\end{tabular}
\end{table}

\paragraph{Results.}
Meta-LSQ-Sobolev reaches a test MSE of $0.019$ (Table~\ref{tab:maml}). The strongest
MAML baseline we trained, with about $8500$ weights, is still $5.5\times$ worse after
training $36\times$ longer, and its error was still falling. The three learned
scalars come out similar for different basis draws (App.~\ref{app:details}).

\paragraph{Controls for the derivative information.}
Meta-LSQ-Sobolev gets support derivatives and MLP-MAML does not, so we ran three
controls. Without derivatives Meta-LSQ is still
$12\times$ better than standard MAML, and MAML given the same derivatives through a
Sobolev loss improves but stays $11\times$ worse than Meta-LSQ-Sobolev. The
ALPaCA-style model is about as accurate as Meta-LSQ, and value-only Meta-LSQ trained as long matches it
in float64 but not in float32 (App.~\ref{app:details}). The derivatives help both methods but do not account for
the gap to MAML.

\paragraph{Comparison with Gaussian processes.}
The closest comparator is GEK, a GP given the same values and derivatives. Fitted
per task it is
slightly more accurate ($0.013$, within seed spread of Meta-LSQ), and a kernel
matched to the sinusoid family solves this benchmark almost exactly, but every task
then needs its own marginal-likelihood search. When we instead share and meta-learn
its three hyperparameters exactly as Meta-LSQ does, the exact GP reaches $0.026$ and
Meta-LSQ is lower on most seeds (App.~\ref{app:gp}). This suggests that most of the
gain over MAML comes from learning one kernel for all tasks and adapting with a
closed-form solve. The random features are at least as accurate here.

\begin{table}[tb]
\centering
\caption{Test MSE after a fixed single-thread wall-clock budget (mean\,$\pm$\,population
std over 3 seeds, all evaluated on the same 200 test tasks). The floored row
applies the ridge floor $\mu \geq 10^{-2}$ of Sec.~\ref{sec:exp-limits} during
meta-training.}
\label{tab:wallclock}
\footnotesize
\setlength{\tabcolsep}{2pt}
\begin{tabular}{lrrr}
\toprule
method & $0.5$\,s & $5$\,s & $50$\,s \\
\midrule
MLP-MAML           & $3.50$\,\tiny$\pm.06$   & $2.07$\,\tiny$\pm.05$   & $0.61$\,\tiny$\pm.06$ \\
Meta-LSQ $w{=}0$   & $.304$\,\tiny$\pm.076$  & $.173$\,\tiny$\pm.106$  & $.045$\,\tiny$\pm.009$ \\
Meta-LSQ-Sob.\     & $\mathbf{.071}$\,\tiny$\pm.017$ & $\mathbf{.034}$\,\tiny$\pm.018$ & $\mathbf{.0076}$\,\tiny$\pm.0011$ \\
\quad $+$ floor    & $.144$\,\tiny$\pm.034$  & $.085$\,\tiny$\pm.033$  & $.020$\,\tiny$\pm.009$ \\
\bottomrule
\end{tabular}
\end{table}

\paragraph{Matched wall-clock budgets.}
Iterations are not comparable units of work, so we also give each method a fixed
window of single-thread training time (0.5, 5 or 50\,s), train it for as many steps
as fit in the window and then test it (Table~\ref{tab:wallclock}). A step costs
about 4\,ms for Meta-LSQ-Sobolev and 5\,ms for MLP-MAML, so Meta-LSQ takes about a
third more steps, and it also gains far more per step. After half a second (about 115 steps) Meta-LSQ-Sobolev is
$8.6\times$ more accurate than MLP-MAML after fifty seconds (about 9100 steps).
Training with the ridge floor costs accuracy here, so elsewhere we apply it only at
deployment (App.~\ref{app:tail}).

\subsection{Real Data: Sensor Calibration under Drift}
\label{sec:exp-real}

\paragraph{Sensor drift.} The response of low-cost metal-oxide gas sensors also
depends on temperature, humidity and ageing, so the curve that converts a reading into a gas concentration drifts and is
re-fitted from a few reference measurements. In the UCI Air Quality
data~\citep{devito2008airquality}, a device with five such sensors in a polluted
urban area logged hourly readings next to a certified reference analyser for about a
year (9357 hours). For one sensor and gas, a week of readings follows a
noisy calibration curve that moves from week to week, so each week is a few-shot
task, with no derivatives measured. Because the drift follows temperature and
humidity, every method except the per-week quadratic may use both as inputs next
to the reading $x$. Meta-LSQ adds
to the sine features $H$ a trend $\Phi$ with rows $(1,\text{inputs},x^2)$ and fits
each week by
\begin{equation}
\label{eq:prior-mean}
\min_{\beta,\gamma}\ \|H\beta+\Phi\gamma-y\|^2+\mu\|\beta\|^2+\mu_\tau\|\gamma-\bar\gamma\|^2,
\end{equation}
where $\bar\gamma$, like $\sigma$, $\mu$ and $\mu_\tau$, is meta-learned across weeks.
With no data the fit is the typical calibration curve $\Phi\bar\gamma$. Ten noisy
points move $\gamma$ only as far as $\mu_\tau$ allows, and the sines add the week's
nonlinear correction. Our first design had no trend, so each week started from zero.
MAML's learned initialisation and ALPaCA's learned prior mean play the role of
$\bar\gamma$, and the GP gets the same trend as its mean.

\paragraph{Setup.} Each ISO week is a task whose calibration curve has to be
recovered from $K{=}10$ hours and is scored on $50$ others. Weeks are split
chronologically (60/40), so the test weeks (November to April) come after all
training weeks, and half of their hours are colder than $95\%$ of the training hours.
We use the four sensor/gas pairs with enough weeks (CO, NO$_2$, benzene and NO$_x$,
53--56 each). We chose the design above after exploratory runs that also saw the test
weeks. Whether each method uses temperature and humidity, and Meta-LSQ's ridge floor,
are then picked by four-fold cross-validation over blocks of training weeks.

\begin{table}[tb]
\centering
\caption{Gas-sensor calibration: mean normalised query MSE on held-out, later weeks
(5 seeds, identical draws for every method). Meta-LSQ and the exact RBF GP have the
learned prior mean and trend, and the inputs of every method except ``quad.'' are
chosen by cross-validation. ``ALPaCA'' is ALPaCA-style with a learned prior mean; ``lin.''
and ``quad.'' are per-week least squares. $^\ast$In units of $10^{-4}$. Bold:
lowest at the printed precision.
Spreads, cost and the design without prior mean in App.~\ref{app:real}.}
\label{tab:real}
\footnotesize
\setlength{\tabcolsep}{2.4pt}
\begin{tabular}{lrrrrrr}
\toprule
target & Meta-LSQ & GP & ALPaCA & MAML & lin. & quad. \\
\midrule
CO      & $0.25$   & $\mathbf{0.20}$    & $\mathbf{0.20}$ & $0.28$   & $0.34$ & $0.40$ \\
NO$_2$  & $0.78$   & $0.69$             & $\mathbf{0.42}$ & $1.61$   & $0.84$ & $1.74$ \\
benzene$^\ast$ & $1.6$ & $\mathbf{0.57}$ & $2.2$ & $1.4$ & $483$ & $3.0$ \\
NO$_x$  & $2.57$   & $\mathbf{0.62}$    & $0.63$          & $0.72$   & $1.56$ & $2.28$ \\
\bottomrule
\end{tabular}
\end{table}

\paragraph{Results.} ALPaCA-style is the most accurate on NO$_2$, the GP on benzene,
and the two are level on CO and NO$_x$ (Table~\ref{tab:real}). Meta-LSQ is below MAML
on every seed for CO and NO$_2$ and level with it on benzene. Our first design
(the reading alone, no prior mean or trend, $N{=}200$ and a floor of $10\mu^\ast$) has
$30\%$, $12\%$ and $25\times$ higher error on these three. On NO$_x$ it fails on three
of five seeds, extrapolating on the colder test weeks. The last cross-validation
block was hard for every method ($1.07$ to $2.14$, Meta-LSQ $1.41$), so it gave no
warning specific to Meta-LSQ. This is roughly what the kernel view of
Sec.~\ref{sec:sobolev-solve} leads us to expect. Without slope rows Meta-LSQ
approximates the shared GP, and with ten points per week the exact kernel is more
accurate and cheaper to train ($1.4$\,s against $5$--$9$\,s, since Meta-LSQ has
$N{=}1000$ features to build; App.~\ref{app:real}). What transfers across weeks is
the typical calibration curve, which the other methods learn as well; Meta-LSQ's
advantages need derivative or operator rows, which these data lack.

\begin{figure}[!b]
\centering
\input{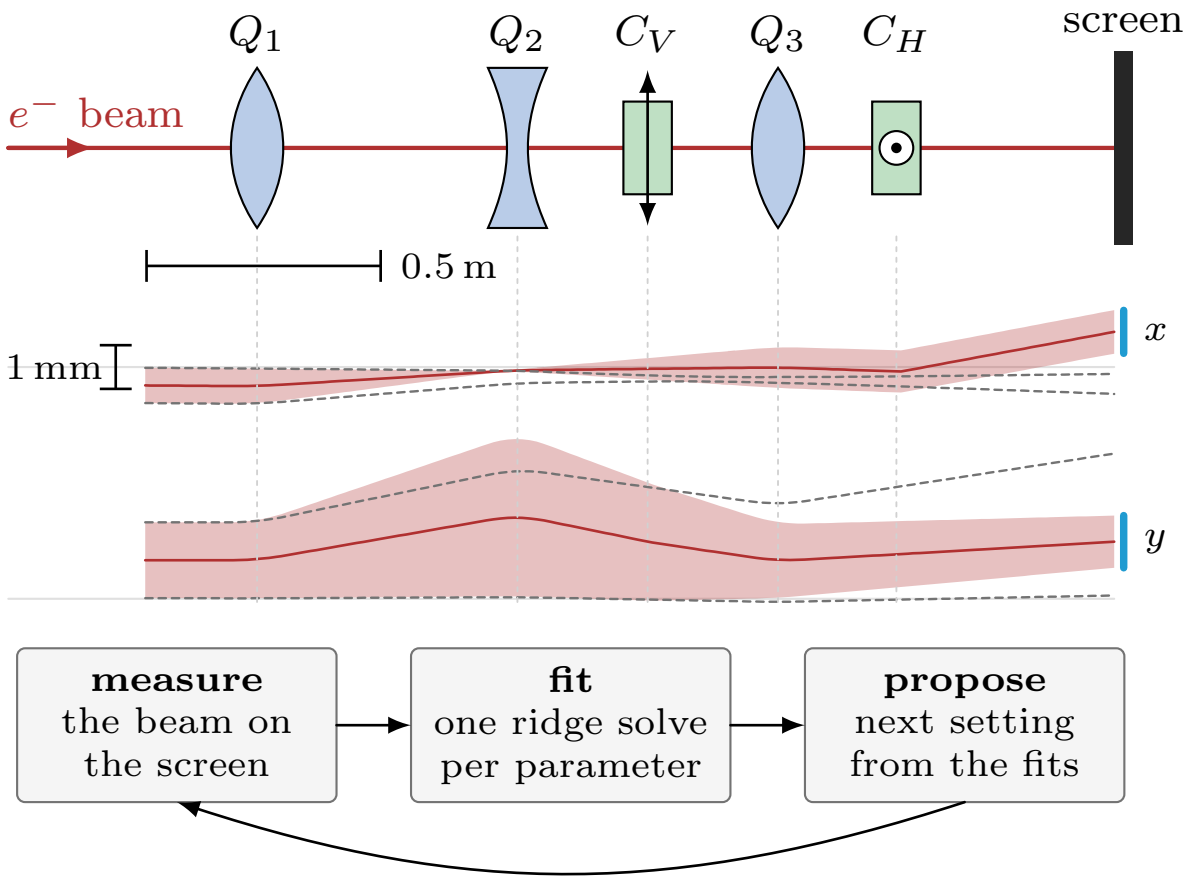}
\caption{The ARES beamline to scale (quadrupoles drawn as lenses that focus or
defocus in $x$, steerers as boxes marked with the plane of their kick and widened),
the beam along it, and the loop of the surrogate methods. The strips show the beam's
centre $\pm2\sigma$ in $x$ and $y$ (magnified, bar 1\,mm) on the trial of
Fig.~\ref{fig:ares-screens}, at the starting setting (grey dashes) and at
Meta-LSQ's tuned setting (red); the cyan bars are the target at the screen.}
\label{fig:ares-scheme}
\end{figure}

\subsection{Closed-Loop Beam Tuning in Simulation}
\label{sec:exp-ares}

\begin{figure*}[tb]
\centering
\includegraphics{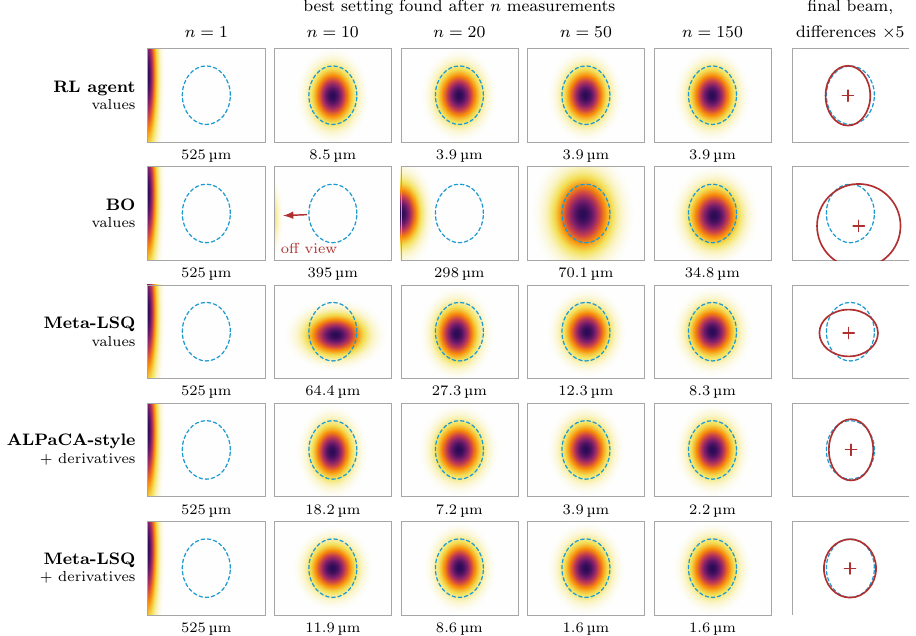}\space\space
\caption{Tuning on the trial where Meta-LSQ with derivatives ends closest to its
median: the simulated screen at the best setting after $n$ measurements, with the
mean absolute error of the four beam parameters; the dashed ellipse is the target
($2\sigma$). Last column: the final beam (solid) against the target, differences
magnified five times.}
\label{fig:ares-screens}
\end{figure*}

We also put the solve inside a tuning loop, on the ARES experimental area at
DESY~\citep{panofski2021ares}. An electron beam passes three quadrupoles, magnetic
lenses that focus it in one plane and defocus it in the other, and two steerers,
small dipoles that deflect it, and then hits a screen
(Fig.~\ref{fig:ares-scheme}). The screen gives four numbers, the horizontal and
vertical position and width of the beam, and tuning means finding the five magnet
strengths that bring them to a requested target in few measurements. The answer
changes from trial to trial with the incoming beam and the misalignments of the
quadrupoles, which the tuner does not know. \citet{kaiser2024rlbo} compared an
optimiser trained by reinforcement learning with Bayesian optimisation on 300 such
trials. We simulate the beamline with Cheetah~\citep{kaiser2024cheetah}, use their
trials, starting setting and budget of 150 measurements, and score the mean
absolute error of the four beam parameters at the best setting found.

\paragraph{Meta-training on simulated machines.} Each beam parameter gets its own
Meta-LSQ model, a function from the five magnet settings (scaled to $[-1,1]$) to that
parameter, with $N{=}500$ features. A meta-training task is one simulated machine,
with incoming beam and misalignments drawn from the ranges of DESY's environment,
measured at 60 settings around a random setting; we use 400 such tasks and none of
the 300 trials. As on the sinusoids, the solve is fitted on a support set (30
settings with derivatives, 50 without) and scored on the rest, and the scalars are
learned by Adam. With derivatives the support rows also hold the slopes and second
derivatives of the beam parameter with respect to the magnets, taken from the
differentiable simulation, which stands in for a model of the real beamline; the
second-derivative rows get a weight of their own, a fourth scalar.

\paragraph{Tuning.} Nothing is trained during a trial. For each of the 150
measurements the loop of Fig.~\ref{fig:ares-scheme} refits the four models to all
measurements so far, one solve each with the learned scalars fixed, and measures
next the setting whose predicted beam is closest to the target, within a small
window around the last setting (App.~\ref{app:ares}). ALPaCA-style models get the
same information in the same loop, each variant was fixed on 60 development tasks,
and DESY's agent and their BoTorch~\citep{balandat2020botorch} optimisation run in
our re-implementation of their protocol. Their optimisation uses measurements only,
so with derivatives Meta-LSQ is compared with ALPaCA-style and with gradient descent.

\begin{table}[tb]
\centering
\caption{Beam tuning on DESY's 300 ARES trials: error of the best setting after 150
measurements (\textmu m) and measurements needed to remove $99\%$ of the starting
error (median, with the share of trials that get there). One training run per
method; protocol in App.~\ref{app:ares}.}
\label{tab:ares}
\footnotesize
\setlength{\tabcolsep}{2.5pt}
\begin{tabular}{llrrr}
\toprule
method & uses & median & mean & to $99\%$ \\
\midrule
RL agent~\citep{kaiser2024rlbo} & values & $\mathbf{4.4}$ & $17.1$ & $39$ ($62\%$) \\
BO~\citep{kaiser2024rlbo}       & values & $30.7$ & $38.0$ & -- ($1\%$) \\
ALPaCA-style                    & values & $7.5$  & $\mathbf{13.2}$ & -- ($46\%$) \\
Meta-LSQ                        & values & $7.8$  & $\mathbf{13.2}$ & -- ($44\%$) \\
\midrule
Gradient descent & $+$ deriv. & $14.2$ & $27.7$ & -- ($37\%$) \\
ALPaCA-style & $+$ deriv. & $1.9$ & $9.8$ & $20$ ($78\%$) \\
Meta-LSQ     & $+$ deriv. & $\mathbf{1.6}$ & $\mathbf{9.6}$ & $\mathbf{16}$ ($79\%$) \\
\bottomrule
\end{tabular}
\end{table}

From measurements alone the two surrogates are level (Table~\ref{tab:ares}) and
both are about four times more accurate than the Bayesian optimisation, while the
agent has the lowest median. Its mean is above the surrogates' because we hold its
magnets to their limits, which DESY's simulation did not do, and without that it is
$11.8$\,\textmu m (App.~\ref{app:ares}). With model
derivatives Meta-LSQ has the lowest error in Table~\ref{tab:ares} and removes $99\%$
of the starting error in 16 measurements, slightly ahead of ALPaCA-style. Gradient descent on the
measured error with the same derivatives ends at $14.2$\,\textmu m, worse than both
surrogates from values alone, so the derivatives help mainly through the fitted
model. Figure~\ref{fig:ares-screens} shows one trial.

\begin{figure*}[tb]
\centering
\includegraphics[width=\textwidth]{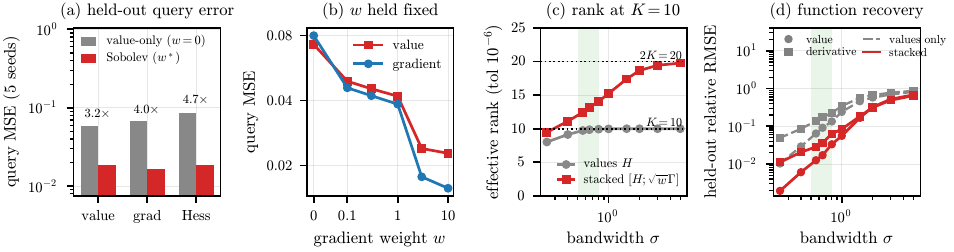}
\caption{Effect of the derivative rows. \textbf{(a)}~On held-out tasks they lower
query value, gradient and Hessian error (5 seeds), though nothing above first order
is supervised. \textbf{(b)}~Meta-training with $w$ held fixed: mean value and
gradient error fall as $w$ grows (3 seeds). \textbf{(c)}~Effective rank at the
10-shot support set (200 draws): the derivative rows add about three directions at
the learned bandwidth (shaded) and approach $2K$ only at large $\sigma$
(Observation~\ref{obs:rank}). \textbf{(d)}~Held-out relative RMSE of an in-span
random target, values-only against stacked fit ($\mu{=}10^{-5}$, mean over 200
draws).}
\label{fig:mechanism}
\end{figure*}

On the first 150 trials we also took the derivatives from an imperfect model, with
no misalignments and the incoming beam off by up to $20\%$ (App.~\ref{app:ares}).
Meta-LSQ and ALPaCA-style then end near $11$\,\textmu m, worse than Meta-LSQ from
values alone ($8.8$\,\textmu m on the same trials) and only slightly ahead of gradient
descent ($12$\,\textmu m). The meta-training tasks have
no readout noise and no magnet hysteresis, so we leave both out of the evaluation.
On a machine each reversal of a magnet also costs time, and the surrogate loop
reverses a quadrupole about 240 times per run against 8 for the agent, so with a
30\,s cycle per reversal the agent removes $99\%$ of the starting error sooner
($4.1$ against $6.6$ minutes, medians; App.~\ref{app:ares}).

DESY's trials ask for beam sizes of at most $0.5$\,mm. On their simulated grid of
targets over the full range (sizes up to $2$\,mm), for which they published the
final beams of the agent and the Bayesian optimisation, the same values-only loops
end at a median of $7.3$\,\textmu m, against $14.4$ and $41.9$. The agent is level with the surrogates below $0.5$\,mm, and its
error grows to $278$\,\textmu m for the largest targets, where the surrogates stay near
$8$\,\textmu m (Table~\ref{tab:ares-grid}). The published agent is scored on the last
beam of its run and the surrogates on their best setting. Scored on its best setting
in our own run, the agent is ahead of Meta-LSQ below $0.5$\,mm and behind it in the
other bands (App.~\ref{app:ares}).

\begin{table}[tb]
\centering
\caption{Simulated tuning on DESY's grid of targets by the larger target beam size:
error after 150 measurements (\textmu m, median / mean), 300 targets per band.
Agent and BO from the final beams DESY published, ALPaCA is the ALPaCA-style
model. The last row weights the bands by their share of the grid.}
\label{tab:ares-grid}
\scriptsize
\setlength{\tabcolsep}{3pt}
\begin{tabular}{lrrrrr}
\toprule
size (mm) & share & agent & BO & ALPaCA & Meta-LSQ \\
\midrule
below $0.5$ & $49\%$ & $6.7/10$   & $29/46$   & $6.5/7.5$ & $6.6/8.9$ \\
$0.5$--$1$  & $23\%$ & $18/152$   & $49/80$   & $7.7/11$  & $7.7/12$ \\
$1$--$1.5$  & $9\%$  & $69/258$   & $109/147$ & $7.9/8.6$ & $7.9/8.4$ \\
$1.5$--$2$  & $19\%$ & $278/1222$ & $189/236$ & $8.3/21$  & $8.3/21$ \\
\midrule
whole grid  &        & $14/295$   & $42/99$   & $7.3/11$  & $7.3/12$ \\
\bottomrule
\end{tabular}
\end{table}

\subsection{Derivative and Operator Rows}
\label{sec:exp-mechanism}

Figure~\ref{fig:mechanism}a and~b repeat the analysis of Sec.~\ref{sec:mechanism} for the
meta-learned solve on held-out tasks. Derivative rows lower the query error of the
values and of every derivative order we checked; the Hessian error falls by
$4.7\times$, although only first derivatives are supervised. With $w$ held fixed
during meta-training, a larger $w$ gives a lower error (App.~\ref{app:details}).

\paragraph{Noisy and missing derivatives.}
As $w = \sigma_y^2/\sigma_g^2$ predicts, noisy derivatives drive the learned $w$
down by more than an order of magnitude, while derivatives that are only missing at some support points leave it
roughly unchanged (App.~\ref{app:wmodel}).

\paragraph{Operator and integral rows.}
Any linear operator $\mathcal{L}$ built from derivatives and integrals maps a
feature $\sin(Wx+b)$ to a closed-form function, so the same solve takes rows
$\mathcal{L}[\sin(Wx+b)]$ with the right-hand side of an equation as targets, plus
rows for the boundary conditions. For
the Helmholtz equation $-u''+k^2u=f$ these rows are $(W^2+k^2)\sin(Wx+b)$, and a
new problem $(k,f)$ is solved by one ridge solve, with no training. The bandwidth
and ridge are meta-learned from the equation residual at held-out points of the
training problems, so no solution is ever used as a label. Against the exact
solution the median error is $1.8\times10^{-7}$, four times lower than choosing the
bandwidth for each problem at a fixed ridge; choosing it per problem at the learned
ridge lowers the error further, to $5.7\times10^{-8}$. Integral rows turn a
Volterra integro-differential equation into one linear system as well, and
gradients through the solve recover its memory parameter from noisy observations to
within $4\%$ (App.~\ref{app:operator}).

\subsection{Out-of-Class Transfer and Failure Modes}
\label{sec:exp-limits}

The sinusoids lie in the span of the basis by construction, so the results above
are a best case. Table~\ref{tab:ooc} tests models meta-trained on sinusoids alone on
two new families: \textbf{tanh}, $y = a\tanh(b(x-\phi))$, which is smooth but
outside the span, and \textbf{triangle}, $y = a\,\mathrm{tri}((x-\phi)/T)$, which
has kinks and any period.

\begin{table}[tb]
\centering
\caption{Out-of-class transfer after meta-training on sinusoids: query MSE per
task, mean and median over 200 test tasks (5 seeds; 3 for $t_2$). The Meta-LSQ-Sobolev
columns use the learned ridge $\mu^\ast$, a floor $\mu\geq10^{-2}$ at test time, and
the floor with a Student-$t$ frequency draw. A few blowups dominate the mean.}
\label{tab:ooc}
\footnotesize
\setlength{\tabcolsep}{1.75pt}
\begin{tabular}{lrrrrrrrr}
\toprule
& \multicolumn{2}{c}{MLP-MAML}
& \multicolumn{2}{c}{$\mu^\ast$}
& \multicolumn{2}{c}{$\mu\!\geq\!10^{-2}$}
& \multicolumn{2}{c}{$t_2$ freq.} \\
\cmidrule(lr){2-3}\cmidrule(lr){4-5}\cmidrule(lr){6-7}\cmidrule(lr){8-9}
family & mean & med & mean & med & mean & med & mean & med \\
\midrule
sin      & $0.70$ & $0.47$ & $\mathbf{.019}$ & $\mathbf{3}$e-5 & $.056$ & $1.7$e-3 & $0.28$ & $.044$ \\
tanh     & $6.0$ & $4.6$ & $45$ & $0.11$ & $0.56$ & $.055$ & $\mathbf{0.18}$ & $\mathbf{.037}$ \\
triangle & $\mathbf{4.0}$ & $3.2$ & $18000$ & $32$ & $63$ & $9.2$ & $14$ & $\mathbf{3.0}$ \\
\bottomrule
\end{tabular}
\end{table}

By the median, Meta-LSQ-Sobolev with its learned ridge beats MLP-MAML on tanh by
about $40\times$, but its mean is far worse. The learned ridge is close to zero, so
on a few unfamiliar tasks the fit goes far off where there are no support points,
and these tasks dominate the mean. A more stable solver does not remove
them, so they come from extrapolation (App.~\ref{app:tail}). A
floor $\mu \geq 10^{-2}$ at test time removes most of them, at a cost on sinusoids
that stays far below MLP-MAML's error, and Meta-LSQ-Sobolev then beats MLP-MAML on tanh in the mean as well.
On triangles MLP-MAML keeps by far the lower mean, and only a
heavier-tailed (Student-$t$) frequency draw brings the median to MLP-MAML's level.

\paragraph{Setting the floor.}
Noise on the support values raises the learned ridge by nearly three orders of
magnitude, as $\mu = \sigma_y^2/\sigma_p^2$ predicts, so a floor amounts to assuming
some noise; with it Meta-LSQ-Sobolev stays well below MLP-MAML at every noise level
(App.~\ref{app:wmodel}). One fixed floor does not suit targets with different noise
levels. On our first real-data design a floor of $10$ to $30$ times the learned
ridge did best on the held-out weeks, and for the final design cross-validation
chose $1$ to $3$ times (App.~\ref{app:real}).

\subsection{Selecting the Ridge}
\label{sec:exp-ridge}

We also tried to choose the ridge from the support data alone, by generalised
cross-validation (GCV) and leave-out cross-validation.

\begin{observation}[In-sample criteria at a vanishing ridge]
\label{obs:ridge}
Suppose the design has fewer rows than features and full row rank, and the support
targets are noiseless. As $\mu \to 0$ the fit passes through every support point,
whatever it does away from them. A criterion built only from the residual at the
support points and its degrees of freedom therefore cannot see how the fit
extrapolates. GCV, for one, tends to a finite value as $\mu \to 0$, so it has no
reason to avoid the smallest ridge (sketch in App.~\ref{app:tail}).
\end{observation}

GCV chose the smallest ridge on every sinusoid task and on most tanh tasks, the best
choice for sinusoids and by far the worst for tanh. Leave-out cross-validation
matches the floor on typical tasks but misses the rare blowups (App.~\ref{app:tail}).
Any kernel or random-feature model fitted to a few noiseless points shares this
limit.

\section{Discussion and Conclusions}
\label{sec:discussion}

We have presented Meta-LSQ, a meta-learner for smooth, low-dimensional few-shot
regression, the setting of our main experiments, that fits each task by one linear
solve. Derivative and operator observations enter that solve as extra rows. It works best when
the task family is close to the span of the basis, and with a ridge floor it also
beats MLP-MAML on smooth out-of-span tanh functions. The comparison with the
shared-hyperparameter GP suggests that most of the improvement over MAML comes from
sharing a three-scalar Sobolev kernel across tasks. In simulated beam tuning the
solve is a competitive surrogate from measurements alone and, given model
derivatives, slightly ahead of ALPaCA-style and well ahead of gradient descent. On real calibration data without derivatives it needed a
learned prior mean to pass MAML, and an exact GP and an ALPaCA-style model stay
more accurate, as expected without slope rows. The ALPaCA-style encoder transfers better out of class
(App.~\ref{app:details}), non-smooth targets remain the main weakness, and derivative
estimates degrade in high dimension (App.~\ref{app:gmm}). With one-hot labels the
solve is a closed-form classifier, so far tested on synthetic tasks only
(App.~\ref{app:classify}). Integrals of the features are closed-form too, so future work
could fit the solve to integral observations, such as readings averaged over a time
window or a quantity known only as its total over a region. On a real beamline the inputs
would also need each magnet's last direction of motion, because of hysteresis.
\label{main:end}

\section*{Acknowledgments}
We thank Innovatrics for supporting this work.

\section*{LLM Disclosure}
We used Claude (Anthropic; Opus 5.5) in Claude Code to write the experiment
code, to help with the methodology, experimental design and interpretation of
results, and to edit the manuscript, including the formal statements and proof
sketches. AI tools were not used for synthetic data, conceptual frameworks,
hypotheses, translation, data cleaning or qualitative data analysis. All code is
publicly available on the author's GitHub,
\url{https://github.com/sulcantonin/metalsq-public}.
\label{disclosure:end}

\printbibliography

\clearpage
\appendix

\section{Training Algorithm}
\label{app:alg}

\begin{algorithm}[tb]
\caption{Meta-LSQ training}
\label{alg:meta-lsq}
\begin{algorithmic}[1]
\Require frozen $\hat W \in \R^{d\times N}$, $\hat W_{kj}\sim\mathcal{N}(0,1)$;
         phases $b_j\sim\mathcal{U}[0,2\pi)$
\Require log-parameters $\ell_\sigma, \ell_\mu, \ell_w$; Adam
\For{outer step $= 1,\dots,T_{\text{outer}}$}
  \State sample $B$ tasks $(X_i,y_i,G_i,X_i^q,y_i^q)$
  \State $\sigma \gets e^{\ell_\sigma}$, $\mu \gets e^{\ell_\mu}$,
         $w \gets e^{\ell_w}$, $W \gets \sigma\hat W$
  \For{$i = 1,\dots,B$}
    \State assemble $H_i,\Gamma_i$ by~\eqref{eq:cyclic}; build $A_i,t_i$
    \State $\beta_i \gets \texttt{solve}(A_i^\top\! A_i + \mu I,\;A_i^\top t_i)$
           \hfill{\footnotesize// differentiable}
    \State $\mathcal{L}_i \gets \|H_i^q\beta_i - y_i^q\|^2/K_q$
  \EndFor
  \State Adam-step on $\nabla_{\ell_\sigma,\ell_\mu,\ell_w}
         \tfrac{1}{B}\sum_i \mathcal{L}_i$
\EndFor
\State \Return $(\sigma^\ast,\mu^\ast,w^\ast,\hat W,b)$
\end{algorithmic}
\end{algorithm}

\section{Posterior, and the Meaning of $\mu$ and $w$}
\label{app:posterior}

Place a Gaussian prior $\beta\sim\mathcal{N}(0,\sigma_p^2 I)$ and independent
Gaussian likelihoods on the two observation types, $y_i = h(x_i)^\top\beta +
\varepsilon^y_i$ with $\varepsilon^y_i \sim \mathcal{N}(0,\sigma_y^2)$ and $g_{ik}
= \gamma_k(x_i)^\top\beta + \varepsilon^g_{ik}$ with $\varepsilon^g_{ik} \sim
\mathcal{N}(0,\sigma_g^2)$. The negative log-posterior is, up to constants,
$\tfrac12\big(\sigma_y^{-2}\|H\beta-y\|^2 + \sigma_g^{-2}\sum_k\|\Gamma_k\beta-g_k\|^2
+ \sigma_p^{-2}\|\beta\|^2\big)$, and multiplying by $2\sigma_y^2$ gives the stacked
objective~\eqref{eq:beta-star} with $w = \sigma_y^2/\sigma_g^2$ and $\mu =
\sigma_y^2/\sigma_p^2$. The posterior is $\beta \mid t \sim \mathcal{N}(\beta^\ast,
\sigma_y^2 (A^\top\! A+\mu I)^{-1})$. Any predictive derivative $D^\alpha
u_N(x_\star) = c^{(\alpha)}(x_\star)^\top\beta$ is linear in $\beta$, so it is
Gaussian with variance $\sigma_y^2\,c^{(\alpha)\top}(A^\top\! A+\mu
I)^{-1}c^{(\alpha)}$, one $\mathcal{O}(N^2)$ product against the cached factor. The
ridge fixes only the ratio $\sigma_y^2/\sigma_p^2$; the absolute scale needs an
estimate of $\sigma_y^2$. Because~\eqref{eq:rff-basis} omits
the factor $\sqrt{2}$, the induced GP has signal variance $\sigma_p^2/2$.

\section{Noise Ablations: the Ridge and $w$}
\label{app:wmodel}

With noise of std $\epsilon \in \{0,0.1,0.25,0.5\}$ on the support values only
(8000 meta-training steps for both methods, 3 seeds), the learned ridge rises from
$1.9\!\times\!10^{-6}$ to $8.6\!\times\!10^{-5}$, $4.2\!\times\!10^{-4}$ and
$1.2\!\times\!10^{-3}$, and Meta-LSQ-Sobolev reaches a clean-query test MSE of
$0.0049$--$0.055$ against about $0.65$ for MLP-MAML. With noise of std
$\epsilon_g \in \{0,0.1,0.25,0.5\}$ on the support gradients only, the learned $w$
falls from $10.9$ to $0.29$, $0.20$ and $0.18$. When only 2, 5 or all 10 support
points carry a clean gradient row (1200 steps, 3 seeds), $w$ stays at $3.8$, $4.9$
and $4.6$.

\section{Gaussian-Process Comparators}
\label{app:gp}

All GPs get the same support values and derivatives as Meta-LSQ-Sobolev (200 tasks,
5 seeds). Fitted per task by marginal likelihood, an RBF kernel reaches $0.013$
($2.9$\,ms per task), and a single spectral-mixture component, whose sample paths
approach $A\sin(\omega x+\phi)$ as its spectral variance goes to zero, reaches $3.2\!\times\!10^{-10}$ ($102$\,ms per
task). The shared GP has an RBF kernel with three hyperparameters (length-scale,
value noise, derivative noise), learned by Adam on the query loss with Meta-LSQ's
tasks, initialisation, learning rate and budget, then frozen. It reaches $0.026$
after 2000 steps and $0.013$ after 6000, against $0.019$ and $0.0043$ for
Meta-LSQ-Sobolev, which is lower on 4 of 5 and on all 5 paired seeds.

\section{Real Data: Protocol and Full Results}
\label{app:real}

Each task is one ISO week of hourly readings, with the reference analyser as target:
PT08.S1 to CO, PT08.S4 to NO$_2$, PT08.S2 to benzene and PT08.S3 to NO$_x$, with
53--56 weeks of at least 60 valid hours (the NMHC reference has only eight weeks).
The first $60\%$ of weeks are for training, and the support (10) and query (50)
hours of a week are a seeded split, identical for every method. Our first design
gave Meta-LSQ the sensor reading as its only input, $N{=}200$ features and
coefficients shrunk toward zero, and the GP the reading alone with a zero mean.

\paragraph{The ridge floor.} Table~\ref{tab:relfloor} applies a floor $\mu =
c\,\mu^\ast$ at test time to the first design. On the noisy targets the error is
flat or improving up to $c \approx 30$; on the almost noiseless benzene it degrades
with $c$, and the fixed floor of Table~\ref{tab:ooc} costs $40\times$.

\begin{table}[tb]
\centering
\caption{Ridge floor on held-out weeks, applied at test time: mean / worst-week
normalised query MSE (5 seeds pooled for the maximum).}
\label{tab:relfloor}
\footnotesize
\setlength{\tabcolsep}{2pt}
\begin{tabular}{lrrrr}
\toprule
floor & CO & NO$_2$ & benzene & NO$_x$ \\
\midrule
$\mu^\ast$          & $.38$ / $5.8$ & $1.03$ / $5.4$ & $.0013$ / $.020$ & $.95$ / $8.0$ \\
$3\mu^\ast$         & $.35$ / $4.1$ & $.92$ / $3.5$  & $.0021$ / $.040$ & $.90$ / $7.3$ \\
$10\mu^\ast$        & $.32$ / $2.4$ & $.86$ / $2.9$  & $.0037$ / $.071$ & $.85$ / $4.3$ \\
$30\mu^\ast$        & $.32$ / $1.6$ & $.84$ / $2.9$  & $.0066$ / $.096$ & $.85$ / $4.6$ \\
$100\mu^\ast$       & $.35$ / $1.5$ & $.89$ / $2.6$  & $.012$ / $.22$   & $.94$ / $4.7$ \\
$\mu \geq 10^{-2}$  & $.32$ / $1.7$ & $.85$ / $2.9$  & $.051$ / $.48$   & $.84$ / $4.3$ \\
\bottomrule
\end{tabular}
\end{table}

\paragraph{Design of Sec.~\ref{sec:exp-real}.} The task set drops hours without a
temperature or humidity reading. Meta-LSQ has $N{=}1000$ features with one bandwidth
per input. With $A = [H,\Phi]$, $\bar\beta = (0,\bar\gamma)$ and $D$ diagonal with
$\mu$ on the sine block and $\mu_\tau$ on the trend, the minimiser
of~\eqref{eq:prior-mean} is $\bar\beta + D^{-1}A^\top(AD^{-1}A^\top + I)^{-1}(y -
A\bar\beta)$, a $10\times10$ solve. The GP has one length-scale per input and the
learned trend as its mean, and ALPaCA-style is the encoder of
Sec.~\ref{sec:exp-works} with a meta-learned prior mean on its last-layer weights.
Cross-validation chose temperature and humidity for Meta-LSQ and the GP except on
benzene, for ALPaCA-style on every target and for MAML only on NO$_2$, and floors
of $3\mu^\ast$, $\mu^\ast$, $3\mu^\ast$ and $3\mu^\ast$. On NO$_x$ Meta-LSQ's test error
per seed is $0.73$, $0.78$, $2.79$, $5.35$ and $3.19$, and $56\%$ of the test hours
are colder than the 5th percentile of the training temperatures. Over the
four targets (one thread) Meta-LSQ has 6--10 meta-learned parameters, trains in
5.4--8.7\,s and adapts to a week in 0.29--0.49\,ms; the GP has 5--9, 1.3--1.4\,s and
0.06--0.07\,ms, ALPaCA-style 1842, 6.9--7.0\,s and 0.09--0.10\,ms, and MAML
1761--1841, 8.8--9.2\,s and 0.12--0.13\,ms. We reached this design in two
exploratory rounds that also scored the test weeks; the numbers come from one run
of the design fixed afterwards, to which ALPaCA-style and the timings were added
later without changing any other row.

\begin{table*}[tb]
\centering
\caption{Gas-sensor calibration on the task set of Sec.~\ref{sec:exp-real}: mean
normalised query MSE on held-out weeks, $\pm$ std over 5 seeds. ``First design''
rows use the sensor reading alone, no prior mean, $N{=}200$ and a floor of
$10\mu^\ast$; the
other rows use the inputs (and, for Meta-LSQ, the floor) chosen by cross-validation.}
\label{tab:realv2}
\footnotesize
\setlength{\tabcolsep}{5pt}
\begin{tabular}{lrrrr}
\toprule
method & CO & NO$_2$ & benzene & NO$_x$ \\
\midrule
Meta-LSQ, first design            & $0.32 \pm 0.05$ & $0.87 \pm 0.08$ & $4.0$e-3 $\pm$ $2.9$e-3 & $0.84 \pm 0.07$ \\
Meta-LSQ, prior mean and trend    & $0.25 \pm 0.03$ & $0.78 \pm 0.08$ & $1.6$e-4 $\pm$ $1.9$e-4 & $2.57 \pm 1.9$ \\
GP RBF, first design              & $0.30 \pm 0.03$ & $0.84 \pm 0.06$ & $3.8$e-3 $\pm$ $2.8$e-3 & $0.81 \pm 0.05$ \\
GP RBF, prior mean and trend      & $0.20 \pm 0.01$ & $0.69 \pm 0.10$ & $5.7$e-5 $\pm$ $3.2$e-5 & $0.62 \pm 0.02$ \\
ALPaCA-style, prior mean          & $0.20 \pm 0.01$ & $0.42 \pm 0.04$ & $2.2$e-4 $\pm$ $1.3$e-4 & $0.63 \pm 0.03$ \\
MLP-MAML                          & $0.28 \pm 0.02$ & $1.61 \pm 0.97$ & $1.4$e-4 $\pm$ $1.2$e-4 & $0.72 \pm 0.03$ \\
per-task linear                   & $0.34 \pm 0.05$ & $0.84 \pm 0.20$ & $4.8$e-2 $\pm$ $0.9$e-2 & $1.56 \pm 0.50$ \\
per-task quadratic                & $0.40 \pm 0.18$ & $1.74 \pm 0.89$ & $3.0$e-4 $\pm$ $2.2$e-4 & $2.28 \pm 1.4$ \\
\bottomrule
\end{tabular}
\end{table*}

\section{Extrapolation and the Ridge}
\label{app:tail}

\paragraph{Sketch of Observation~\ref{obs:ridge}.} Write $A = U\,\mathrm{diag}(s)\,V^\top$,
$c = U^\top t$ and $\rho_i = \mu/(s_i^2+\mu)$. The in-sample residual is $\sum_i
\rho_i^2 c_i^2$ and the residual degrees of freedom are $\sum_i \rho_i$. Full row
rank means every $s_i > 0$, so both vanish as $\mu\to0$, while GCV $= n\sum_i
\rho_i^2 c_i^2/(\sum_i \rho_i)^2$, with $n$ rows, tends to $n\sum_i c_i^2
s_i^{-4}/(\sum_i s_i^{-2})^2$.

\paragraph{Solver accuracy and the tail.} A stabilised least-squares solve on
$[A;\sqrt{\mu}I]$ instead of the normal equations changes the in-class test MSE by
less than $1\%$ and the blowup ratio $\max|\text{prediction}|/\max|\text{target}|$
by at most $3.2$ where it reaches $980$, so the tail comes from extrapolation. Raising
$\mu$ cuts it (Table~\ref{tab:tail}). Leave-out cross-validation over $\mu$ (7
support points to fit, 3 to score) and GCV are compared with the floor in
Table~\ref{tab:ridgepolicy}. With $\mu$ pinned at $10^{-2}$ during meta-training,
one of three runs diverged by $2\!\times\!10^4$ steps.

\begin{table}[tb]
\centering
\caption{Per-task MSE on the triangle family across ridge values (600 tasks pooled
over 3 seeds).}
\label{tab:tail}
\small
\setlength{\tabcolsep}{4pt}
\begin{tabular}{lrrrrr}
\toprule
ridge $\mu$ & median & p90 & p99 & max & top-1\% \\
\midrule
$\mu^\ast\!\approx\!10^{-5}$ & $34$ & $3900$ & $2.6$e5 & $4.8$e6 & $76\%$ \\
$10^{-3}$ & $12$ & $282$ & $4400$ & $1.7$e4 & $43\%$ \\
$10^{-2}$ & $7.7$ & $124$ & $762$ & $3800$ & $31\%$ \\
$10^{-1}$ & $\mathbf{7.1}$ & $\mathbf{74}$ & $\mathbf{182}$ & $\mathbf{582}$ & $\mathbf{15\%}$ \\
\bottomrule
\end{tabular}
\end{table}

\begin{table}[tb]
\centering
\caption{Ridge policies on 1000 pooled per-task fits per family (5 seeds $\times$
200 tasks, float64). This run is separate from the one behind Table~\ref{tab:ooc}
and pools its medians, so the two differ slightly.}
\label{tab:ridgepolicy}
\footnotesize
\setlength{\tabcolsep}{3pt}
\begin{tabular}{llrrr}
\toprule
family & policy & mean & median & p99 \\
\midrule
\multirow{4}{*}{sin}
 & tuned $\mu^\ast$ & $0.019$ & $2.7$e-5 & $0.39$ \\
 & floor $10^{-2}$   & $0.054$ & $1.5$e-3 & $1.0$ \\
 & leave-out CV      & $0.024$ & $5.2$e-5 & $0.47$ \\
 & GCV               & $\mathbf{0.0062}$ & $\mathbf{7.8}$\textbf{e-7} & $\mathbf{0.091}$ \\
\midrule
\multirow{4}{*}{tanh}
 & tuned $\mu^\ast$ & $44$          & $0.100$ & $1070$ \\
 & floor $10^{-2}$   & $\mathbf{0.57}$ & $\mathbf{0.057}$ & $\mathbf{8.8}$ \\
 & leave-out CV      & $12$          & $0.065$ & $211$ \\
 & GCV               & $1948$        & $0.227$ & $4.2$e4 \\
\midrule
\multirow{4}{*}{triangle}
 & tuned $\mu^\ast$ & $1.8$e4       & $27.8$ & $2.6$e5 \\
 & floor $10^{-2}$   & $\mathbf{63}$ & $8.8$  & $\mathbf{914}$ \\
 & leave-out CV      & $5419$        & $\mathbf{7.1}$ & $7.4$e4 \\
 & GCV               & $1.5$e6       & $54.5$ & $3.2$e7 \\
\bottomrule
\end{tabular}
\end{table}

\section{Operator and Integral Rows}
\label{app:operator}

\paragraph{Helmholtz.}
\label{app:pde}
For $-u''+k^2u=f$ on $[0,2\pi]$ with homogeneous Dirichlet conditions,
$k\sim\mathcal{U}[1,10]$ and $f$ a random sum of low-frequency sinusoids, the rows
are $(W^2+k^2)\sin(Wx+b)$, and $(\sigma,\mu)$ are meta-learned from the PDE residual
at held-out collocation points. Scored against the exact solution (5 seeds), the
shared pair reaches a median error of $1.8\!\times\!10^{-7}$ against
$8.0\!\times\!10^{-7}$ for choosing among seven bandwidths per task at ridge
$10^{-6}$; given the learned ridge, per-task selection reaches
$5.7\!\times\!10^{-8}$.

\paragraph{Volterra.}
\label{app:integro}
Integrals of the sine features are also closed-form, so
\begin{equation}
\label{eq:volterra}
u'(x) + \lambda\!\int_0^x\! u(s)\,ds = f(x),\quad u(0)=0,\; x\in[0,1],
\end{equation}
is one linear system. Given $f$ and noisy observations of $u$ at 60 points, Adam
through the solve recovers $\lambda \in \{0.5,1,2,4\}$ to within $0.3$--$3.9\%$
under $5\%$ observation noise.

\section{Few-Shot Classification}
\label{app:classify}

One-hot labels turn~\eqref{eq:beta-star} into a closed-form one-vs-rest classifier.
On synthetic 5-way Gaussian-blob tasks in $d{=}2$ (3 seeds) it reaches $89.0\%$
(5-shot) and $84.0\%$ (1-shot), against $85.1\%$ and $80.5\%$ for MLP-MAML.

\section{Log-Density Fits and Dimension}
\label{app:gmm}

Fitting the log-density of a Gaussian mixture from samples with values and gradients
($N{=}1500$, bandwidth selected on held-out data, one seed), the value error stays at
or below $6\!\times\!10^{-2}$ for $d$ up to 64, while gradient and Hessian errors
reach $0.89$ and $0.96$ at $d{=}64$.

\section{Posterior Calibration}
\label{app:calib}

For value-only Meta-LSQ with support noise of $0$, $0.1$ and $0.25$ (300 tasks, 3
seeds), nominal $90\%$ intervals with the residual plug-in $\hat\sigma_y^2 =
\frac1M\|H\beta^\ast-y\|^2$ cover $31$--$64\%$; $\sigma_y^2{=}\mu$ covers $27$--$68\%$
and $\sigma_y^2{=}1$ about $100\%$.

\section{Beam Tuning on ARES}
\label{app:ares}

\paragraph{Protocol.} We rebuilt the simulated ARES section of DESY's rl-vs-bo code
in Cheetah (Gaussian beam moments, no noise, an unbounded screen). The five magnet
settings are scaled to $[-1,1]$, every method starts at $(10,-10,0,10,0)$ (in m$^{-2}$ and mrad), and each
move stays within $\pm5\%$ of a magnet's range around the last setting, as in
DESY's Bayesian optimisation. A surrogate step fits one model per beam parameter to
all measurements so far, predicts the beam at 2000 random settings in that window
and measures the one predicted closest to the target (refined by gradient descent
with derivatives; from values alone the predicted log error is lowered by $0.1$
times the range of the measured log errors times the posterior standard deviation,
normalised to at most 1, to favour unexplored settings). The surrogates, gradient descent and the Bayesian
optimisation keep each quadrupole's polarity fixed (strengths from $0$ to $\pm72$\,m$^{-2}$
in our re-implementation), whereas the agent may reverse it, and its best setting
does so on 110 of the 300 trials. Meta-LSQ uses $N{=}500$ and 1500 Adam steps, ALPaCA-style two tanh
layers of 40 and 8000 steps; both are meta-trained on 400 simulated tasks of 60
settings. Each model's variant (values, slopes, second derivatives) was chosen on 60
development tasks. Gradient descent steps from the best setting along the
normalised gradient of the distance to the target, halving the step when a
measurement does not improve. DESY's agent runs from its published weights; our copy
repeats the moves of their published runs at every step and their errors (median
$4.4$, mean $11.3$\,\textmu m). Their simulation lets the magnets leave their limits,
which the agent does on 28 of 300 trials; holding the magnets to the limits raises
its mean from $11.8$ to $17.1$\,\textmu m and changes its median little ($4.3$ to
$4.4$\,\textmu m).

\paragraph{Limits.} Each method was trained once. The ALPaCA-style architecture was
not tuned, whereas Meta-LSQ's settings come from earlier development on this task,
and the development tasks were drawn from ranges chosen after looking at DESY's
trials.

\paragraph{Imperfect model.} The imperfect model of Sec.~\ref{sec:exp-ares} has no
misalignments and an incoming beam whose sizes and divergences are off by up to
$20\%$, position by $0.2$\,mm, angle by $20$\,\textmu rad and energy by $2\%$; the
readings come from the true machine and only the derivatives from the model.

\paragraph{Machine time.} With a 1\,s readout, a 10\,s ramp over a magnet's full
range, 1\,s settling and a 30\,s cycle per reversal to clear hysteresis, the agent
removes $95\%$ of the starting error in $2.2$ minutes and Meta-LSQ with derivatives
in $3.3$, and $99\%$ in $4.1$ and $6.6$ (medians); without the cycle Meta-LSQ needs
$0.6$ minutes for $99\%$.

\paragraph{Targets over the full range.} DESY's grid has $160{,}000$ targets
(positions within $\pm2$\,mm, sizes $20$\,\textmu m to $2$\,mm, one incoming beam).
We drew 300 from each band of Table~\ref{tab:ares-grid}. Run by us and scored on its
best setting, the agent reaches $5.5$, $15$, $34$ and $185$\,\textmu m, ahead of
Meta-LSQ in the first band and behind it on $82\%$, $95\%$ and $97\%$ of the targets
in the others.

\section{Protocol and Further Results}
\label{app:details}

\paragraph{Protocol.} All experiments run single-threaded on CPU; the rows of
Table~\ref{tab:maml} come from two machines, so their times are indicative. A seed
redraws the frozen basis and the training and test tasks. The outer loop starts at
$\sigma{=}1$, $\mu{=}10^{-4}$, $w{=}1$.

\paragraph{Baselines and training.} The standard MLP-MAML has two hidden layers of 40
ReLU units and one inner step at $10^{-2}$ (Adam at $10^{-3}$, 8000 iterations). The
strong one is first-order MAML with three hidden layers of 64, 5 inner steps at
$5\!\times\!10^{-3}$ and gradient clipping. Meta-LSQ has $N{=}200$ features, Adam at
$5\!\times\!10^{-3}$ for 2000 iterations on batches of 4 tasks, and
Meta-LSQ-Sobolev gets the analytic gradients $g_i = a\cos(x_i+\phi)$. The test MSE
is $0.024$, $0.021$ and $0.023$ for $N \in \{50,100,200\}$ (3 seeds). A solve takes $0.81$\,ms
and an outer step $9.6$\,ms against $8.1$\,ms for MLP-MAML. The learned scalars are
$\sigma = 0.65 \pm 0.09$, $\mu^\ast \approx 10^{-5}$ and $w = 10.9 \pm 3.4$ over
3 basis draws.

\paragraph{Controls.} The ALPaCA-style model (two tanh layers of 40, a closed-form
Bayesian last layer, 8000 steps on values) reaches $0.026$ against $0.023$ for
Meta-LSQ-Sobolev on matched seeds, and value-only Meta-LSQ trained as long reaches
$0.019$ in float64 and $0.041$ in float32, the precision of our other runs. Out of class, ALPaCA-style gives $0.30$ on tanh against $0.64$
for floored Meta-LSQ.

\paragraph{Derivative rows.} At $\sigma{=}0.65$ (200 draws, tolerance $10^{-6}$) the
effective rank is $9.9$ with values and $13.1$ of 20 with slopes, and the ratio of
the largest to the tenth singular value drops from $10^{4.7}$ to $10^{2.3}$. The
coefficient error is about as large as the coefficients in both designs, while the
held-out error of the fit falls from $0.091$ to $0.017$ for values and from $0.17$
to $0.037$ for derivatives. With $w$ fixed in meta-training at
$\{0,0.1,0.3,1,3,10\}$ (3 seeds), the error falls from $0.073$ to $0.023$ for values,
though not on every seed.

\paragraph{Bandwidth matrix and spectrum.} On $y = a\sin(\omega_1x_1+\omega_2x_2+\phi)$
with $\omega_1 \sim \mathcal{U}[0.5,1.5]$ and $\omega_2 \sim \mathcal{U}[3,5]$, a
diagonal bandwidth matrix is $3.9\times$ better than a scalar and a triangular one a
further $1.8\times$ (5 of 5 seeds). With the floor, a Student-$t_2$ frequency draw
cuts the triangle median from $7.99$ to $2.97$ (MLP-MAML: $3.15$) but raises the
in-class median from $0.0015$ to $0.044$ (3 seeds).

\end{document}